\documentclass[10pt,twocolumn,letterpaper]{article}

\usepackage{iccv}
\usepackage{times}
\usepackage{epsfig}
\usepackage{graphicx}
\usepackage{amsmath}
\usepackage{amssymb}
\usepackage{multirow}
\usepackage{booktabs}
\usepackage{eqnarray}

\usepackage{makecell}

\usepackage{pifont}% http://ctan.org/pkg/pifont

\usepackage{comment}

\usepackage{color}

\usepackage[pagebackref=true,breaklinks=true,letterpaper=true,colorlinks,bookmarks=false]{hyperref}

\iccvfinalcopy % *** Uncomment this line for the final submission

\def\iccvPaperID{****} % *** Enter the ICCV Paper ID here

\begin{document}

%%%%%%%%% TITLE
\title{RepPoints: Point Set Representation for Object Detection}

\author{Ze Yang$^{1\dag}$\thanks{Equal contribution. $^\dag$The work is done when Ze Yang and Shaohui Liu are interns at Microsoft Research Asia.} \quad Shaohui Liu$^{2,3\dag*}$ \quad Han Hu$^3$ \quad Liwei Wang$^1$ \quad Stephen Lin$^{3}$\\
    $^1$Peking University \quad 
    $^2$Tsinghua University\\
	$^3$Microsoft Research Asia \\
	{\tt\small yangze@pku.edu.cn, b1ueber2y@gmail.com, wanglw@cis.pku.edu.cn} \\
	{\tt\small\{hanhu, stevelin\}@microsoft.com} \\
}

\maketitle
%\thispagestyle{empty}

%%%%%%%%% ABSTRACT
\begin{abstract}
   
Modern object detectors rely heavily on rectangular bounding boxes, such as anchors, proposals and the final predictions, to represent objects at various recognition stages. The bounding box is convenient to use but provides only a coarse localization of objects and leads to a correspondingly coarse extraction of object features. In this paper, we present \textbf{RepPoints} (representative points), a new finer representation of objects as a set of sample points useful for both localization and recognition. Given ground truth localization and recognition targets for training, RepPoints learn to automatically arrange themselves in a manner that bounds the spatial extent of an object and indicates semantically significant local areas. They furthermore do not require the use of anchors to sample a space of bounding boxes. We show that an anchor-free object detector based on RepPoints can be as effective as the state-of-the-art anchor-based detection methods, with 46.5 AP and 67.4 $AP_{50}$ on the COCO test-dev detection benchmark, using ResNet-101 model. Code is available at \href{https://github.com/microsoft/RepPoints}{\color{cyan}{https://github.com/microsoft/RepPoints}}.

\end{abstract}

%%%%%%%%% BODY TEXT
\section{Introduction}
Object detection aims to localize objects in an image and provide their class labels. As one of the most fundamental tasks in computer vision, it serves as a key component for many vision applications, including instance segmentation \cite{pinheiro2015learning}, human pose analysis \cite{toshev2014deeppose}, and visual reasoning \cite{wu2017learning}. The significance of the object detection problem together with the rapid development of deep neural networks has led to substantial progress in recent years \cite{erhan2014scalable,girshick2014rich,girshick2015fast,ren2015faster,he2015spatial,dai2016r}. 

\begin{figure}[tb]
\centering
\vspace{-10pt}
\includegraphics[width = 0.95\linewidth]{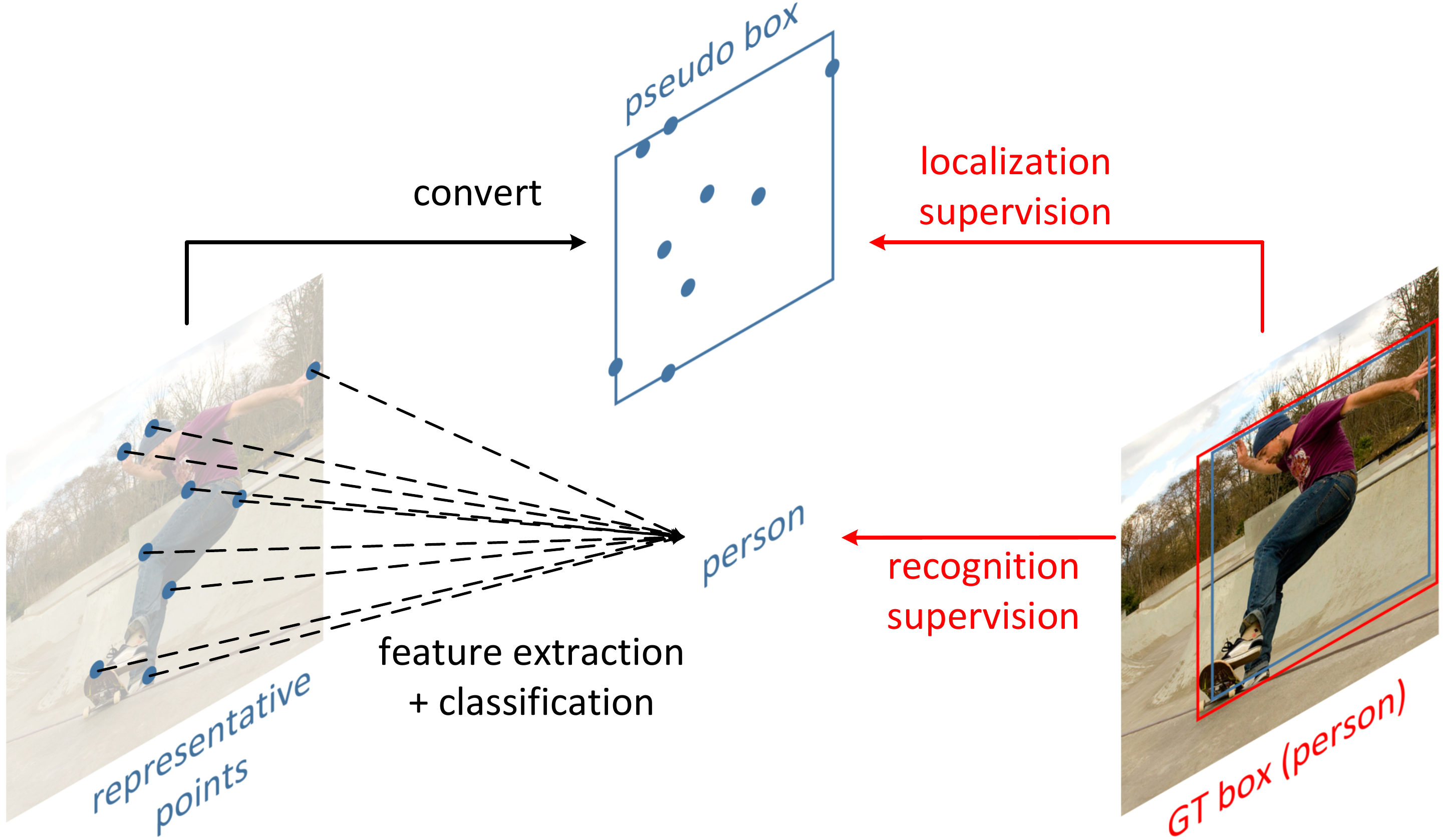}
\caption{\textit{RepPoints} is a new representation for object detection that consists of a set of points which indicate the spatial extent of an object and semantically significant local areas. This representation is learned via weak localization supervision from rectangular ground-truth boxes and implicit recognition feedback. Based on the richer RepPoints representation, we develop an anchor-free object detector that yields improved performance compared to using bounding boxes.}
\vspace{-10pt}
\label{fig::demo}
\end{figure}

In the object detection pipeline, bounding boxes, which encompass rectangular areas of an image, serve as the basic element for processing. They describe target locations of objects throughout the stages of an object detector, from anchors and proposals to final predictions. Based on these bounding boxes, features are extracted and used for purposes such as object classification and location refinement. The prevalence of the bounding box representation can partly be attributed to common metrics for object detection performance, which account for the overlap between estimated and ground truth bounding boxes of objects. Another reason lies in its convenience for feature extraction in deep networks, because of its regular shape and the ease of subdividing a rectangular window into a matrix of pooled cells.

Though bounding boxes facilitate computation, they provide only a coarse localization of objects that does not conform to an object's shape and pose. Features extracted from the regular cells of a bounding box may thus be heavily influenced by background content or uninformative foreground areas that contain little semantic information. This may result in lower feature quality that degrades classification performance in object detection.

In this paper, we propose a new representation, called \textit{RepPoints}, that provides more fine-grained localization and facilitates classification. Illustrated in Fig.~\ref{fig::demo}, RepPoints is a set of points that learns to adaptively position themselves over an object in a manner that circumscribes the object's spatial extent and indicates semantically significant local areas. The training of RepPoints is driven jointly by object localization and recognition targets, such that the RepPoints are tightly bound by the ground-truth bounding box and guide the detector toward correct object classification. This adaptive and differentiable representation can be coherently used across the different stages of a modern object detector, and does not require the use of anchors to sample over a space of bounding boxes.

RepPoints differs from existing non-rectangular representations for object detection, which are all built in a bottom-up manner~\cite{Denet,CornerNet,ExtremeNet}. These bottom-up representations identify individual points (e.g., bounding box corners or object extremities) and rely on handcrafted clustering to group them into object models. Their representations furthermore either are still axis-aligned like bounding boxes \cite{Denet,CornerNet} or require ground truth object masks as additional supervision \cite{ExtremeNet}. In contrast, RepPoints are learned in a top-down fashion from the input image / object features, allowing for end-to-end training and producing fine-grained localization without additional supervision. 

With RepPoints replacing all the conventional bounding box representations in a two-stage object detector, including anchors, proposals and final localization targets, we develop a clean and effective \emph{anchor-free} object detector, which achieves 42.8 AP and 65.0 $AP_{50}$ on the COCO benchmark \cite{MSCOCO} without multi-scale training and testing, and achieves 46.5 AP and 67.4 $AP_{50}$ with multi-scale training and testing using ResNet-101 model. The proposed object detector not only surpasses all existing anchor-free detectors but also performing on-par with state-of-the-art anchor-based baselines.

\section{Related Work}
\label{sec::related-work}
\paragraph{Bounding boxes for the object detection problem.}
The bounding box has long been the dominant form of object representation in the field of object detection. One reason for its prevalence is that a bounding box is convenient to annotate with little ambiguity, while providing sufficiently accurate localization for the subsequent recognition process. This may explain why the major benchmarks all utilize annotations and evaluations based on bounding boxes \cite{PascalVOC,MSCOCO,OpenImagesV4}. In turn, these benchmarks motivate object detection methods to use the bounding box as their basic representation in order to align with the evaluation protocols.

Another reason for the dominance of bounding boxes is that almost all image feature extractors, both before \cite{Haar,HOG} and during the deep learning era \cite{AlexNet,VGG,GoogLeNet,ResNet}, are based on an input patch with a regular grid form. It is thus convenient to use the bounding box representation to facilitate feature extraction \cite{girshick2014rich,girshick2015fast,ren2015faster}.

Although the proposed \textit{RepPoints} has an irregular form, we show that it can be amenable to convenient feature extraction. Our system utilizes RepPoints in conjunction with deformable convolution~\cite{DCN}, which naturally aligns with RepPoints in that it aggregates information from input features at several sample points. Besides, a rectangular \textit{pseudo box} can be readily generated from RepPoints (see Section \ref{sec::reppoints}), allowing the new representation to be used with object detection benchmarks. 

\vspace{-10pt}
\paragraph{Bounding boxes in modern object detectors.}
The best-performing object detectors to date generally follow a multi-stage recognition paradigm \cite{FPN,DCN,Mask-rcnn,li2017light,Cascade-rcnn,SNIP,PA-Net}, and the bounding box representation appears in almost all stages: 1) as pre-defined~\cite{ren2015faster,RetinaNet} or learnt~\cite{GuidedAnchoring,MetaAnchor,zhong2018anchor} anchors that serve as hypotheses over the bounding box space; 2) as refined object proposals connecting successive recognition stages~\cite{ren2015faster,Cascade-rcnn}; and 3) as the final localization targets.

The RepPoints can be used to replace the bounding box representations in all stages of model object detectors, resulting in a more effective new object detector. Specifically, the anchors are replaced by center points, which is a special configuration of RepPoints. The bounding box proposals and final localization targets are replaced by the RepPoints proposals and final targets. Note the resulting object detector is anchor-free, due to the use of center points for initial object representation. It is thus even more convenient in use than the bounding box based methods.

\vspace{-10pt}
\paragraph{Other representations for object detection.}

To address limitations of rectangular bounding boxes, there have been some attempts to develop more flexible object representations. These include an elliptic representation for pedestrian detection \cite{leibe2005pedestrian} and a rotated bounding box to better handle rotational variations \cite{huang2007high,zhou2017oriented}.

Other works aim to represent an object in a bottom-up manner. Early bottom-up representations include DPM \cite{felzenszwalb2010object} and Poselet \cite{bourdev2010detecting}. Recently, bottom-up approaches to object detection have been explored with deep networks \cite{CornerNet,ExtremeNet}. CornerNet \cite{CornerNet} first predicts top-left and bottom-right corners and then employs a specialized grouping method \cite{newell2017associative} to obtain the bounding boxes of objects. However, the two opposing corner points still essentially model a rectangular bounding box. ExtremeNet~\cite{ExtremeNet} is proposed to locate the extreme points of objects in the x- and y-directions \cite{papadopoulos2017extreme} with supervision from ground-truth mask annotations. In general, bottom-up detectors benefit from a smaller hypothesis space (for example, CornerNet and ExtremeNet both detect 2-d points instead of directly detecting a 4-d bounding box) and potentially finer-grained localization. However, they have limitations such as relying on handcrafted clustering or post-processing steps to compose whole objects from the detected points.

Similar to these bottom-up works, \textit{RepPoints} is also a flexible object representation. However, the representation is constructed in a top-down manner, without the need for handcrafted clustering steps. RepPoints can automatically learn extreme points and key semantic points without supervision beyond ground-truth bounding boxes, unlike ExtremeNet \cite{ExtremeNet} where additional mask supervision is required.

\vspace{-10pt}
\paragraph{Deformation modeling in object recognition.} One of the most fundamental challenges for visual recognition is to recognize objects with various geometric variations. To effectively model such variations, a possible solution is to make use of bottom-up composition of low-level components. Representative detectors along this direction include DPM \cite{felzenszwalb2010object} and Poselet \cite{bourdev2010detecting}. An alternative is to implicitly model the transformations in a top-down manner, where a lightweight neural network block is applied on input features, either globally \cite{STN} or locally \cite{DCN}.

\textit{RepPoints} is inspired by these works, especially the top-down deformation modeling approach \cite{DCN}. The main difference is that we aim at developing a flexible object representation for accurate \emph{geometric localization} in addition to semantic feature extraction. 
In contrast, both the deformable convolution and deformable RoI pooling methods are designed to improve feature extraction only. The inability of deformable RoI pooling to learn accurate geometric localization is examined in Section~\ref{sec::framework} and appendix.
In this sense, we expand the usage of adaptive sample points in previous geometric modeling methods \cite{STN,DCN} to include finer localization of objects.

\vspace{-.5em}
\section{The RepPoints Representation}
We first review the bounding box representation and its use within multi-stage object detectors. This is followed by a description of RepPoints and its differences from bounding boxes.

\subsection{Bounding Box Representation}

\label{sec::bbox}

The bounding box is a 4-d representation encoding the spatial location of an object, $\mathcal{B}=(x, y, w, h)$, with $x, y$ denoting the center points and $w, h$ denoting the width and height. Due to its simplicity and convenience in use, modern object detectors heavily rely on bounding boxes for representing objects at various stages of the detection pipeline.

\vspace{-5pt}
\paragraph{Review of Multi-Stage Object Detectors} The best performing object detectors usually follow a multi-stage recognition paradigm, where object localization is refined stage by stage. The role of the object representation through the steps of this pipeline is as follows:
\begin{equation}
\begin{tabular}{rll}
bbox anchors & $\xrightarrow{\text{bbox reg.}}$ & bbox proposals (S1) \\
       & $\xrightarrow{\text{bbox reg.}}$ &  bbox proposals (S2) \\
       & ...&  \\
       & $\xrightarrow{\text{bbox reg.}}$ & bbox object targets \\
\end{tabular}
\label{eq.bbox_evolution}
\end{equation}

At the beginning, multiple anchors are hypothesized to cover a range of bounding box scales and aspect ratios. In general, high coverage is obtained through dense anchors over the large 4-d hypothesis space. For instance, 45 anchors per location are utilized in RetinaNet \cite{RetinaNet}. 

For an anchor, the image feature at its center point is adopted as the object feature, which is then used to produce a confidence score about whether the anchor is an object or not, as well as the refined bounding box by a bounding box regression process. The refined bounding box is denoted as ``bbox proposals (S1)''.

In the second stage, a refined object feature is extracted from the refined bounding box proposal, usually by RoI-pooling \cite{girshick2015fast} or RoI-Align \cite{Mask-rcnn}. For the two-stage framework \cite{ren2015faster}, the refined feature will produce the final bounding box target by bounding box regression. For the multi-stage approach \cite{Cascade-rcnn}, the refined feature is used to generate intermediate refined bounding box proposals (S2), also by bounding box regression. This step can be iterated multiple times before producing the final bounding box target.

In this framework, bounding box regression plays a central role in progressively refining object localization and object features. We formulate the process of bounding box regression in the following paragraph.

\vspace{-5pt}
\paragraph{Bounding Box Regression} 
Conventionally, a 4-d regression vector $(\Delta x_p, \Delta y_p, \Delta w_p, \Delta h_p)$ is predicted to map the current bounding box proposal $\mathcal{B}_p=(x_p, y_p, w_p, h_p)$ into a refined bounding box $\mathcal{B}_r$, where
\begin{equation}
\mathcal{B}_r = (x_p+w_p \Delta x_p, y_p+h_p \Delta y_p, w_p e^{\Delta w_p}, h_p e^{\Delta h_p}).
\end{equation}
 
Given the ground truth bounding box of an object $\mathcal{B}_t=(x_t, y_t, w_t, h_t)$, the goal of bounding box regression is to have $\mathcal{B}_r$ and $\mathcal{B}_t$ as close as possible. Specifically, in the training of an object detector, we use the distance between the predicted 4-d regression vector and the expected 4-d regression vector  $\hat{\mathcal{F}}(\mathcal{B}_p, \mathcal{B}_t)$ as the learning target, using a smooth $l_1$ loss: 
 \begin{equation}
    \hat{\mathcal{F}}(\mathcal{B}_p, \mathcal{B}_t)=(\frac{x_t - x_p}{w_p}, \frac{y_t - y_p}{h_p}, \log\frac{w_t}{w_p}, \log\frac{h_t}{h_p}).
    \label{eq::bbox-regression}
\end{equation}

This bounding box regression process is widely used in existing object detection methods. It performs well in practice when the required refinement is small, but it tends to perform poorly when there is large distance between the initial representation and the target. Another issue lies in the scale difference between $\Delta x, \Delta y$ and $\Delta w, \Delta h$, which requires tuning of their loss weights for optimal performance.

\subsection{RepPoints}
\label{sec::reppoints}
As previously discussed, the 4-d bounding box is a coarse representation of object location. The bounding box representation considers only the rectangular spatial scope of an object, and does not account for shape and pose and the positions of semantically important local areas, which could be used toward finer localization and better object feature extraction. 

To overcome the above limitations, \textit{RepPoints} instead models a set of adaptive sample points:
\begin{equation}
    \mathcal{R} = \{(x_k, y_k)\}_{k=1}^{n},
\end{equation}
where $n$ is the total number of sample points used in the representation. In our work, $n$ is set to 9 by default.

\vspace{-5pt}
\paragraph{RepPoints refinement} Progressively refining the bounding box localization and feature extraction is important for the success of multi-stage object detection methods. For RepPoints, the refinement can be expressed simply as
\begin{equation}
\label{eq::dbox_refine}
    \mathcal{R}_r = \{(x_k + \Delta x_k, y_k + \Delta y_k)\}_{k=1}^{n},
\end{equation}
where $\{(\Delta x_k, \Delta y_k)\}_{k=1}^{n}$ are the predicted offsets of the new sample points with respect to the old ones. We note that this refinement does not face the problem of scale differences among the bounding box regression parameters, since the offsets are at the same scale in the refinement process of \textit{RepPoints}.

\begin{figure*}[tb]
\centering
 \includegraphics[width=0.9\linewidth]{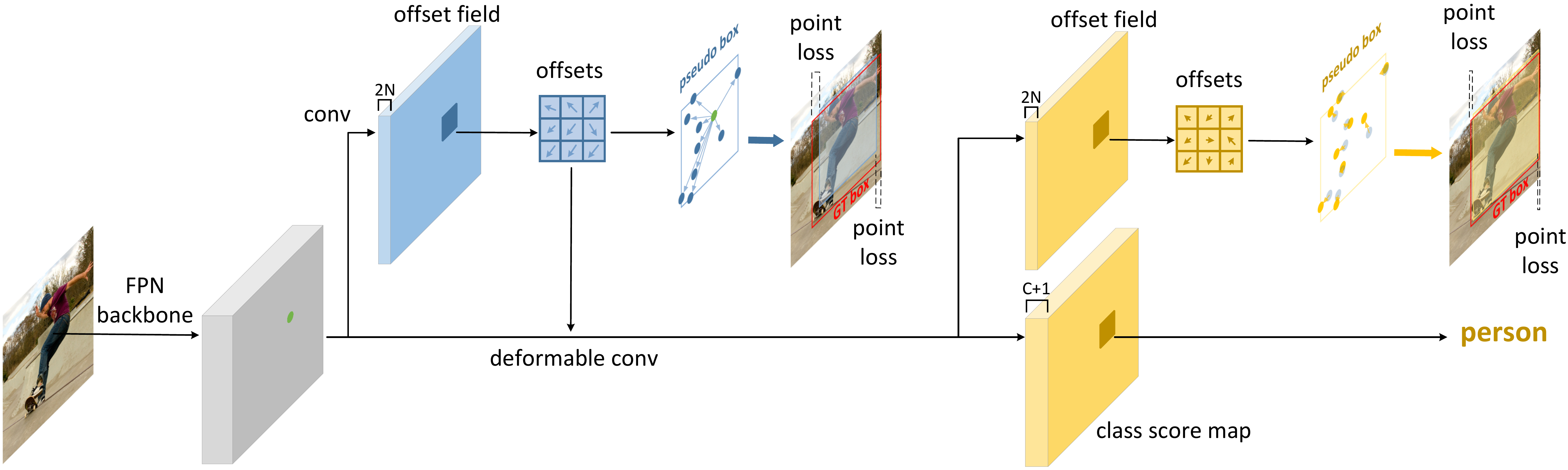}
\caption{Overview of the proposed RPDet (RepPoints detector). While feature pyramidal networks (FPN) \cite{FPN} are adopted as the backbone, we only draw the afterwards pipeline of one scale of FPN feature maps for clear illustration. Note all scales of FPN feature maps share the same afterwards network architecture and the same model weights.}
\vspace{-1em}
\label{fig::pipeline}
\end{figure*}

\vspace{-10pt}
\paragraph{Converting RepPoints to bounding box}
\label{sec::convert}
To take advantage of bounding box annotations in the training of RepPoints, as well as to evaluate RepPoint-based object detectors, a method is needed for converting RepPoints into a bounding box. We perform this conversion using a pre-defined converting function $\mathcal{T}: \mathcal{R}_P \rightarrow \mathcal{B}_P$, where $\mathcal{R}_P$ denotes the RepPoints for object $P$ and $\mathcal{T}(\mathcal{R}_P)$ represents a \textit{pseudo box}.

Three converting functions are considered for this purpose: 
\begin{itemize}
\item \textbf{$\mathcal{T}=\mathcal{T}_1$: Min-max function.} Min-max operation over both axes are performed over the \textit{RepPoints} to determine $\mathcal{B}_p$, equivalent to the bounding box over the sample points.
\item \textbf{$\mathcal{T}=\mathcal{T}_2$: Partial min-max function.} Min-max operation over a subset of the sample points is performed over both axes to obtain the rectangular box $\mathcal{B}_p$.
\item \textbf{$\mathcal{T}=\mathcal{T}_3$: Moment-based function.} The mean value and the standard deviation of the RepPoints is used to compute the center point and scale of the rectangular box $\mathcal{B}_p$, where the scale is multiplied by globally-shared learnable multipliers $\lambda_x$ and $\lambda_y$.
\end{itemize}
These functions are all differentiable, enabling end-to-end learning when inserted into an object detection system. In our experiments, we found them to work comparably well.

\vspace{-10pt}
\paragraph{Learning RepPoints}

The learning of RepPoints is driven by both an object localization loss and an object recognition loss. To compute the object localization loss, we first convert \textit{RepPoints} into a \textit{pseudo box} using the previously discussed transformation function $\mathcal{T}$. Then, the difference between the converted \textit{pseudo box} and the ground-truth bounding box is computed. In our system, we use the smooth $l_1$ distance between the top-left and bottom-right points to represent the localization loss. This smooth $l_1$ distance does not require the tuning of different loss weights as done in computing the distance between bounding box regression vectors (i.e., for $\Delta x, \Delta y$ and $\Delta w, \Delta h$).
Figure~\ref{fig::visualization} indicates that when the training is driven by this combination of object localization and object recognition losses, the extreme points and semantic key points of objects are automatically learned ($\mathcal{T}_1$ is used in transforming RepPoints to pseudo box).

\vspace{-.5em}
\section{RPDet: an Anchor Free Detector}
\label{sec::framework}

We design an anchor-free object detector that utilizes RepPoints in place of bounding boxes as its basic representation. Within a multi-stage pipeline, the object representation evolves as follows:
\begin{equation}
\begin{tabular}{rll}
object centers & $\xrightarrow{\text{RP refine}}$ & RepPoints proposals (S1) \\
       & $\xrightarrow{\text{RP refine}}$ &  RepPoints proposals (S2) \\
       & ...&  \\
       & $\xrightarrow{\text{RP refine}}$ & RepPoints object targets \\
\end{tabular}
\label{eq.dbox_evolution}
\end{equation}

Our RepPoints Detector (RPDet) is constructed with two recognition stages based on deformable convolution, as illustrated in Figure~\ref{fig::pipeline}. Deformable convolution pairs nicely with RepPoints, as its convolutions are computed on an irregularly distributed set of sample points and conversely its recognition feedback can guide training for the positioning of these points. In this section, we present the design of RPDet and discuss its relationship to and differences from existing object detectors.

\vspace{-5pt}
\paragraph{Center point based initial object representation.}

While predefined anchors dominate the representation of objects in the initial stage of object detection, we follow YOLO \cite{YOLO} and DenseBox~\cite{DenseBox} by using center points as the initial representation of objects, which leads to an anchor-free object detector.

An important benefit of the center point representation lies in its much tighter hypothesis space compared to the anchor based counterparts. While anchor based approaches usually rely on a large number of multi-ratio and multi-scale anchors to ensure dense coverage of the large 4-d bounding box hypothesis space, a center point based approach can more easily cover its 2-d space. In fact, all objects will have center points located within the image.

However, the center point based method also faces the problem of recognition target ambiguity, caused by two different objects locating at the same position in a feature map, which limits its prevalence in modern object detectors. In previous methods \cite{YOLO}, this is mainly addressed by producing multiple targets at each position, which faces another issue of vesting ambiguity\footnote{If the center points of multiple ground truth objects are located at a same feature map position, only one randomly chosen ground truth object is assigned to be the target of this position.}. In RPDet, we show that this issue can be greatly alleviated by using the FPN structure \cite{FPN} for the following reasons: first, objects of different scales will be assigned to different image feature levels, which addresses objects of different scales and the same center points locations; second, FPN has a high-resolution feature map for small objects, which also reduces the chance of two objects having centers located at the same feature position. In fact, we observe that only 1.1\% of objects in the COCO datasets \cite{MSCOCO} suffer from the issue of center points located at the same position when FPN is used.

It is worth noting that the center point representation can be viewed as a special RepPoints configuration, where only a single fixed sample point is used, thus maintaining a coherent representation throughout the proposed detection framework.

\vspace{-5pt}
\paragraph{Utilization of RepPoints.} As shown in Figure~\ref{fig::pipeline}, RepPoints serve as the basic object representation throughout our detection system. Starting from the center points, the first set of \textit{RepPoints} is obtained via regressing offsets over the center points. The learning of these \textit{RepPoints} is driven by two objectives: 1) the top-left and bottom-right points distance loss between the induced \textit{pseudo box} and the ground-truth bounding box; 2) the object recognition loss of the subsequent stage. As illustrated in Figure~\ref{fig::visualization}, extreme and key points are automatically learned. The second set of \emph{RepPoints} represents the final object localization, which is refined from the first set of \emph{RepPoints} by Eq.~(\ref{eq::dbox_refine}). Driven by the points distance loss alone, this second set of \textit{RepPoints} aims to learn finer object localization.

\begin{figure}[tb]
\centering
 \includegraphics[width=0.95\linewidth]{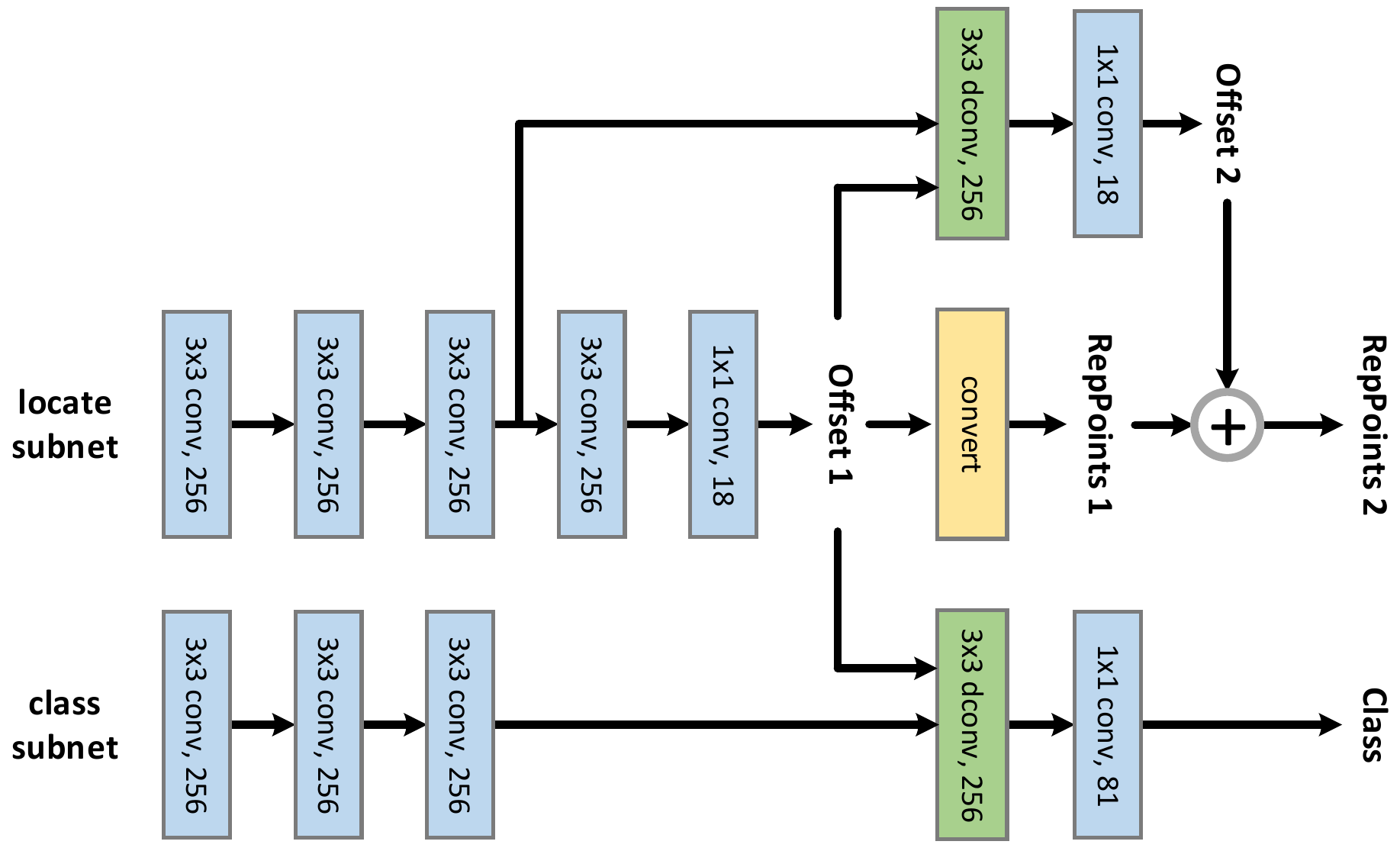}
\caption{The head architecture of RPDet.}
\vspace{-1em}
\label{fig::head_arch}
\end{figure}

\vspace{-10pt}
\paragraph{Relation to deformable RoI pooling~\cite{DCN}.} As mentioned in Section \ref{sec::related-work}, deformable RoI pooling plays a different role in object detection compared to the proposed \emph{RepPoints}. Basically, \textit{RepPoints} is a geometric representation of objects, reflecting more accurate semantic localization, while deformable RoI pooling is geared towards learning stronger appearance features of objects. In fact, deformable RoI pooling cannot learn sample points representing accurate localization of objects (please see the appendix for a proof).

We also note that deformable RoI pooling can be complementary to RepPoints, as indicated in Table~\ref{tab::dpool}.

\vspace{-5pt}
\paragraph{Backbone and head architectures}

Our FPN backbone follows~\cite{RetinaNet}, which produces 5 feature pyramid levels from stage 3 (downsampling ratio of 8) to stage 7 (downsampling ratio of 128). 

The head architecture is illustrated in Figure~\ref{fig::head_arch}. There are two \emph{non-shared} subnets, aiming at localization (RepPoints generation) and classification, respectively. The localization subnet first apply three 256-d $3\times3$ conv layers, followed by two successive small networks to compute offsets for the two sets of \emph{RepPoints}. The classification subnetwork also apply three 256-d $3\times3$ conv layers, followed by a 256-d $3\times3$ deformable conv layer with its input offset field shared with that of the first deformable conv layer in the localization subnetwork. The group normalization layer is applied after each of the first three 256-d $3\times 3$ conv layers in both subnets. 

Note although our approach utilizes two stages of localization, it is even more efficient than the one stage RetinaNet~\cite{RetinaNet} detector (210.9 vs. 234.5 GFLOPS using ResNet-50). The
additional localization stage introduces little overhead due to layer sharing. The anchor-free design reduces the burden of the final classification layer which leads to a slight reduction in computation.

\vspace{-5pt}
\paragraph{Localization/class target assignment.} 
There are two localization stages: generating the first set of RepPoints by refining from the object center point hypothesis (feature map bins); generating the second set of RepPoints by refining from the first RepPoints set. For both stages, only \emph{positive} object hypothesis are assigned with localization (RepPoints) targets in training. For the first localization stage, a feature map bin is \emph{positive} if 1) the pyramidal level of this feature map bin equals the log scale of a ground-truth object $s(B)=\lfloor log_2(\sqrt{w_Bh_B} / 4) \rfloor$; 2) the projection of this ground-truth object's center point locate within this feature map bin. For the second localization stage, the first RepPoints is \emph{positive} if its induced pseudo box has sufficient overlap with a ground-truth object that their intersection-over-union is larger than 0.5.

Classification is conducted on the first set of RepPoints only. The classification assignment criterion follows~\cite{RetinaNet} that IoU (between the induced pseudo box and ground-truth bounding box) larger than 0.5 indicates \emph{positive}, smaller than 0.4 indicates background, and otherwise ignored. Focal loss is employed for classification training~\cite{RetinaNet}.

\vspace{-.5em}
\section{Experiments}

\subsection{Experimental Settings}
We present experimental results of our proposed RPDet framework on the MS-COCO \cite{MSCOCO} detection benchmark, which contains 118k images for training, 5k images for validation (\texttt{minival}) and 20k images for testing (\texttt{test-dev}). All the ablation studies are conducted on \texttt{minival} with ResNet-50 \cite{ResNet}, if not otherwise specified. The state-of-the-art comparison is reported on \texttt{test-dev} in Table \ref{table::system}.

\begin{table}[tb]
	\begin{center}
	\begin{tabular}{c|c|c|c|c}
	\hline
		Representation & Backbone & $AP$ & $AP_{50}$ & $AP_{75}$ \\
	\hline
	\hline
	Bounding box  & ResNet-50 & 36.2 & 57.3 & 39.8 \\
	\hline
	RepPoints (ours) & ResNet-50 & \textbf{38.3} & \textbf{60.0} & \textbf{41.1} \\
	\hline
	\hline
	Bounding box  & ResNet-101 & 38.4 & 59.9 & 42.4 \\
	\hline
	RepPoints (ours) & ResNet-101 & \textbf{40.4} & \textbf{62.0} & \textbf{43.6} \\
	\hline
	\end{tabular}
	\end{center}
\caption{Comparison of the RepPoints and bounding box representations in object detection. The network structures are the same except for processing the given object representation.}
\label{tab::reg-vs-sample-pts}
\vspace{-.5em}
\end{table}

\begin{figure*}[tb]
\centering
\includegraphics[width = 0.9\linewidth]{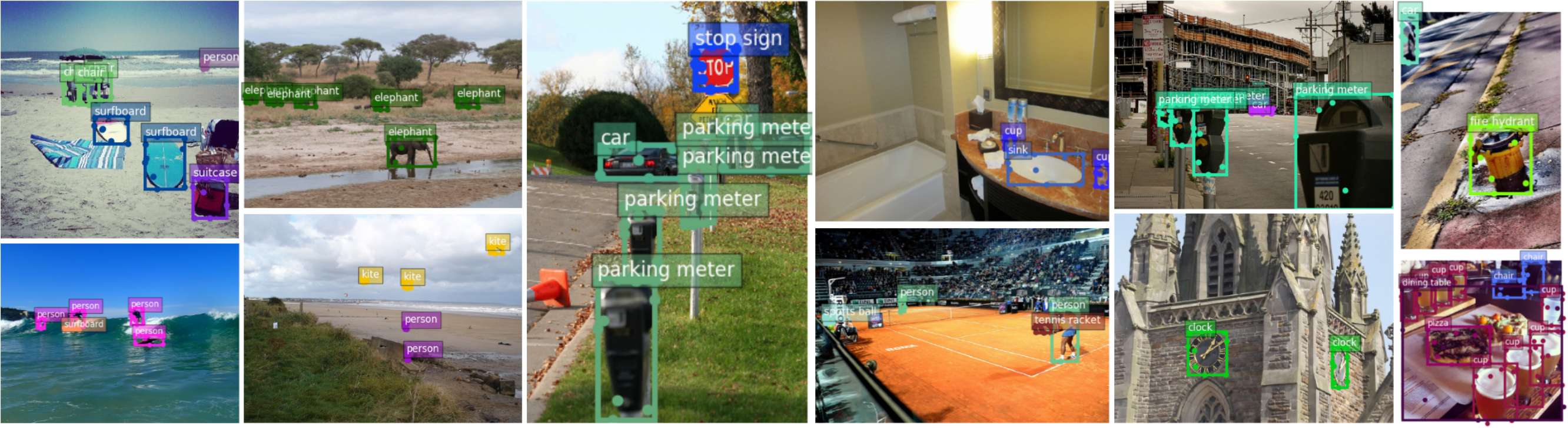}
\caption{Visualization of the learnt RepPoints and the induced bounding boxes on several examples from the COCO~\cite{MSCOCO} minival set (using the pseudo box converting function $\mathcal{T}_1$). In general, the learned RepPoints are located on extreme or semantic keypoints of objects.}
\label{fig::visualization}
\end{figure*}

\begin{table}[tb]
	\begin{center}
	\begin{tabular}{c|c|c|c|c|c}
	\hline
	\multirow{2}{*}{\makecell{Representation}} & \multicolumn{2}{c|}{Supervision}  & \multirow{2}{*}{$AP$} & \multirow{2}{*}{$AP_{50}$} & \multirow{2}{*}{$AP_{75}$} \\
	\cline{2-3}
	& loc. & rec. & & & \\
	\hline
	\hline
	\multirow{2}{*}{bounding box} & \checkmark & & 36.2 & 57.3 & 39.8 \\
	\cline{2-6}
	 & \checkmark &\checkmark& 36.2 & 57.5 & 39.8 \\
	\hline
	\multirow{3}{*}{RepPoints} & & \checkmark & 33.8 & 54.3 & 35.8 \\
	\cline{2-6}
	 & \checkmark & & 37.6 & 59.4 & 40.4 \\
	\cline{2-6}
	 & \checkmark & \checkmark & \textbf{38.3} & \textbf{60.0} & \textbf{41.1} \\
	\hline
    \end{tabular}
	\end{center}
\caption{Ablation of the supervision sources, for both bounding box and RepPoints based object detection. ``loc.'' indicates the object localization loss. ``rec.'' indicates the object recognition loss from the next detection stage.}
\label{tab::reppoints_learning}
\end{table}

\begin{table}[tb]
	\begin{center}
	\begin{tabular}{c|c|c|c|c}
	\hline
	other repr. & init repr. & $AP$ & $AP_{50}$ & $AP_{75}$ \\
	\hline
	\multirow{2}{*}{bounding box} & single anchor  & 36.2 & 57.3 & 39.8 \\
	\cline{2-5}
	& center point & \textbf{37.3} & \textbf{58.0} & \textbf{40.0} \\
	\hline
	\multirow{2}{*}{RepPoints} & single anchor & 36.9 & 58.2 & 39.7 \\
 	\cline{2-5}
    & center point & \textbf{38.3} & \textbf{60.0} & \textbf{41.1} \\ 
	\hline
	\end{tabular}
	\end{center}
\caption{Comparison of using a single anchor and a center point as the initial object representation (``init repr.''). ``other repr.'' indicates representation method for object proposals and final targets.}
\label{tab::loader}
\end{table}

\begin{table}[tb]
	\begin{center}
	\begin{tabular}{c|c|c|c}
	\hline
	method & backbone &\makecell{\# anchors\\per scale} & AP \\
	\hline
	\hline
	RetinaNet \cite{RetinaNet} & ResNet-50 & $3\times3$ & 35.7 \\
	\hline
	FPN-RoIAlign \cite{FPN} & ResNet-50 & $3\times1$ & 36.7 \\
	\hline
	YOLO-like & ResNet-50 & - & 33.9 \\
	\hline
	RPDet (ours) & ResNet-50 & - & \textbf{38.3} \\
	\hline
	\hline
	RetinaNet \cite{RetinaNet} & ResNet-101 & $3\times3$ & 37.8 \\
	\hline
	FPN-RoIAlign \cite{FPN} & ResNet-101 & $3\times1$ & 39.4 \\
	\hline
	YOLO-like & ResNet-101 & - & 36.3 \\
	\hline
	RPDet (ours) & ResNet-101 & - & \textbf{40.4} \\
	\hline
	\end{tabular}
	\end{center}
\caption{Comparison of the proposed method (RPDet) with anchor-based methods (RetinaNet, FPN-RoIAlign) and an anchor-free method (YOLO-like). The YOLO-like method is adapted from the YOLOv1 method~\cite{YOLO} by additionally introducing FPN \cite{FPN}, GN \cite{GN} and focal loss~\cite{RetinaNet} for better accuracy.}
\label{tab::anchors}
\end{table}

Our detector is trained with synchronized stochastic gradient descent (SGD) over 4 GPUs with a total of 8 images per minibatch (2 images per GPU). The ImageNet \cite{deng2009imagenet} pretrained model was used for initialization. Our learning rate schedule follows the `1x' setting \cite{Detectron2018}. Random horizontal image flipping is adopted in training. In inference, NMS ($\sigma=0.5$) is employed to post-process the results, following \cite{RetinaNet}.

\subsection{Ablation Study}

\paragraph{RepPoints vs.~bounding box.}
To demonstrate the effectiveness of the proposed \textit{RepPoints}, we compare the proposed RPDet to a baseline detector where RepPoints are all replaced by the regular bounding box representation.

\noindent \emph{Baseline detector based on bounding box representations.} A single anchor with scale of 4 and aspect ratio of $1:1$ is adopted for the initial object representation, where the anchor box is \emph{positive} if the IoU with a ground-truth object is larger than 0.5. The two sets of RepPoints are replaced by bounding box representation, where the geometric refinement is achieved by the standard bounding box regression method, and the feature extraction is replaced by the RoIAlign~\cite{Mask-rcnn} method using $3\times 3$ grid points\footnote{It can be also implemented by a deformable convolution operator with an unlearnable input offset field induced by the $3\times 3$ grid points.}. All other settings are the same as in the proposed RPDet method.

Table \ref{tab::reg-vs-sample-pts} shows the comparison of two detectors. While the bounding box based method achieves 36.2 mAP, indicating a strong baseline detector, the change of object representation from bounding box to RepPoints brings a +2.1 mAP improvement using ResNet-50~\cite{ResNet} and a +2.0 mAP improvement using a ResNet-101 \cite{ResNet} backbone, demonstrating the advantage of the RepPoints representation over bounding boxes for object detection.

\begin{table}[tb]
	\begin{center}
	\begin{tabular}{c|c|c|c}
	\hline
	\makecell{pseudo box\\converting function} & $AP$ & $AP_{50}$ & $AP_{75}$ \\
	\hline
	    $\mathcal{T}=\mathcal{T}_1$: min-max & 38.2 & 59.7 & 40.7 \\
 	\hline
 		$\mathcal{T}=\mathcal{T}_2$: partial min-max & 38.1 & 59.6 & 40.5 \\
 	\hline
 		$\mathcal{T}=\mathcal{T}_3$: moment-based & 38.3 & 60.0 & 41.1 \\
 	\hline
	\end{tabular}
	\end{center}
\caption{Comparison of different transformation functions from RepPoints to pseudo box, $\mathcal{T}$.}
\label{tab::function}
\end{table}
\begin{table}[tb]
	\begin{center}
	\begin{tabular}{c|c|c|c|c}
	\hline
	\makecell{representation\\method} & w. dpool & $AP$ & $AP_{50}$ & $AP_{75}$ \\
	\hline
	\hline
	\multirow{2}{*}{bounding box} & & 36.2 & 57.3 & 39.8 \\
	\cline{2-5}
 	 & \checkmark & 36.9 & 58.0 & 41.0 \\
	\hline
	\multirow{2}{*}{RepPoints} & & 38.3 & 60.0 & 41.1 \\
	\cline{2-5}
	 & \checkmark & \textbf{39.1} & \textbf{60.6} & \textbf{42.4} \\
	\hline
    \end{tabular}
	\end{center}
\caption{The effect of applying the deformable RoI pooling layer~\cite{DCN} on the proposals of the first stages (see Eq.~(\ref{eq.bbox_evolution}) and Eq.~(\ref{eq.dbox_evolution})). The deformable RoI pooling layer can boost both the methods using bounding boxes and RepPoints, respectively.}
\label{tab::dpool}
\end{table}

\begin{table*}
\begin{center}
\begin{tabular}{l@{\ \ }c@{\ \ \ \ \ }c@{\ \ \ \ \ }c@{\ }c@{\ }c@{\ \ \ \ \ }c@{\ }c@{\ }c}
\toprule
 & Backbone & Anchor-Free & $AP$ & $AP_{50}$ & $AP_{75}$ & $AP_{S}$ & $AP_{M}$ & $AP_{L}$ \\
\midrule
        Faster R-CNN w. FPN \cite{FPN} & ResNet-101 & & 36.2 & 59.1 & 39.0 & 18.2 & 39.0 & 48.2\\
        RefineDet \cite{RefineDet} & ResNet-101 & & 36.4 & 57.5 & 39.5 & 16.6 & 39.9 & 51.4 \\
        RetinaNet \cite{RetinaNet} & ResNet-101 & & 39.1 & 59.1 & 42.3
        & 21.8 & 42.7 & 50.2 \\
        Deep Regionlets \cite{xu2018deep} & ResNet-101 & & 39.3 & 59.8 & - & 21.7
        & 43.7 & 50.9\\
        Mask R-CNN \cite{Mask-rcnn} & ResNeXt-101 & & 39.8 & 62.3 & 43.4 &
        22.1 & 43.2 & 51.2 \\
        FSAF \cite{zhu2019feature} & ResNet-101 & & 40.9 & 61.5 & 44.0 & 24.0 & 44.2 & 51.3 \\
        Cascade R-CNN \cite{Cascade-rcnn} & ResNet-101 & & 42.8 & 62.1 &
        46.3 & 23.7 & 45.5 & 55.2\\
        \midrule
        CornerNet \cite{CornerNet} & Hourglass-104 & \checkmark & 40.5 & 56.5 & 43.1 & 19.4 & 42.7 & 53.9 \\
        ExtremeNet \cite{ExtremeNet}  & Hourglass-104 & \checkmark & 40.1 & 55.3 & 43.2 & 20.3 & 43.2 & 53.1 \\
        \textbf{RPDet} & ResNet-101 & \checkmark & 41.0 & 62.9 & 44.3 & 23.6 & 44.1 & 51.7 \\
        \textbf{RPDet} & ResNet-101-DCN & \checkmark & \textbf{42.8} & \textbf{65.0} & \textbf{46.3} & \textbf{24.9} & \textbf{46.2} & \textbf{54.7} \\
        \textbf{RPDet (ms train)} & ResNet-101-DCN & \checkmark & \textbf{45.0} & \textbf{66.1} & \textbf{49.0} & \textbf{26.6} & \textbf{48.6} & \textbf{57.5} \\
        \textbf{RPDet (ms train \& ms test)} & ResNet-101-DCN & \checkmark & \textbf{46.5} & \textbf{67.4} & \textbf{50.9} & \textbf{30.3} & \textbf{49.7} & \textbf{57.1} \\
\bottomrule
\end{tabular}
\end{center}
\caption{Comparison of the proposed RPDet with the previous state-of-the-art detectors on COCO \cite{MSCOCO} \texttt{test-dev}. The proposed RPDet can achieve 46.5 AP, significantly surpasses all other detectors. Also note the $AP_{50}$ of RPDet is relatively high, which is believed as a better metric in many applications~\cite{YOLOv3}. ``ms'' indicates multi-scale.}
\label{table::system}
\end{table*}

\vspace{-5pt}
\paragraph{Supervision source for RepPoints learning.}
RPDet uses both an object localization loss and an object recognition loss (gradient from later recognition) to drive the learning of the first set of RepPoints, which represents the object proposals of the first stage. Table \ref{tab::reppoints_learning} ablates the use of these two supervision sources in the learning of the object representations. As mentioned before, describing the geometric localization of objects is an important duty of a representation method. Without the object localization loss, it is hard for a representation method to accomplish this duty, as it results in significant performance degradation of the object detectors. For RepPoints, we observe a huge drop of 4.5 mAP by removing the object localization supervision, showing the importance of describing the geometric localization for an object representation method.

Table \ref{tab::reppoints_learning} also demonstrates the benefit of inluding the object recognition loss in learning RepPoints (+0.7 mAP). The use of the object recognition loss can drive the RepPoints to locate themselves at semantically meaningful positions on an object, which leads to fine-grained localization and improves object feature extraction for the following recognition stage. Note that the object recognition feedback cannot benefit object detection with the bounding box representation (see the first block in Table~\ref{tab::reppoints_learning}), further demonstrating the advantage of RepPoints in flexible object representation.

\vspace{-5pt}
\paragraph{Anchor-free vs. anchor-based.}
We first compare the center point based method (a special RepPoints configuration) and the prevalent anchor based method in representing initial object hypotheses, in Table~\ref{tab::loader}. For both detectors using bounding boxes and RepPoints, the center point based method surpass the anchor based method by +1.1 mAP and +1.4 mAP, respectively, likely because of its better coverage of ground-truth objects.

We also compare the proposed anchor-free detector based on RepPoints to RetinaNet~\cite{RetinaNet} (a popular one-stage anchor-based method), FPN~\cite{FPN} with RoIAlign (a popular two-stage anchor-based method) \cite{Mask-rcnn}, and a YOLO-like detector which is adapted from the anchor-free method of YOLOv1~\cite{YOLO}, in Table~\ref{tab::anchors}. The proposed method outperforms both RetinaNet \cite{RetinaNet} and the FPN \cite{FPN} methods, which utilize multiple anchors per scale and sophisticated anchor configurations (FPN). The proposed method also significantly surpasses another anchor-free method (the YOLO-like detector) by +4.4 mAP and +4.1 mAP, respectively, probably due to the flexible RepPoints representation and its effective refinement.

\vspace{-5pt}
\paragraph{Converting RepPoints to pseudo box.}
Table \ref{tab::function} shows that different instantiations of the transformation functions $\mathcal{T}$ presented in Section \ref{sec::reppoints} work comparably well.

\vspace{-5pt}
\paragraph{RepPoints act complementary to deformable RoI pooling \cite{DCN}.}
Table \ref{tab::dpool} shows the effect of applying the deformable RoI pooling layer \cite{DCN} to both bounding box proposals and RepPoints proposals. While applying the deformable RoI pooling layer to bounding box proposals brings a +0.7 mAP gain, applying it to RepPoints proposals also brings a +0.8 mAP gain, implying that the roles of deformable RoI pooling and the proposed RepPoints are complementary.

\subsection{RepPoints Visualization}
In Figure \ref{fig::visualization}, we visualize the learned \textit{RepPoints} and the corresponding detection results on several examples from the COCO \cite{MSCOCO} \texttt{minival} set. It can be observed that \textit{RepPoints} tend to be located at extreme points or key semantic points of objects. These point distributed over objects are automatically learned without explicit supervision. The visualized results also indicate that the proposed RPDet, implemented here with the min-max transformation function, can effectively detect tiny objects.

\subsection{State-of-the-art Comparison}
We compare RPDet with state-of-the-art detectors on \texttt{test-dev}. Table \ref{table::system} shows the results. Without any bells and whistles, RPDet achieves 42.8 AP on COCO benchmark \cite{MSCOCO}, which is on-par with 4-stage anchor-based Cascade R-CNN \cite{Cascade-rcnn} and outperforms all existing anchor-free approaches. By adopting multi-scale training and testing (see Appendix for details), RPDet can achieve 46.5 AP, significantly surpassing all other detectors. Also note the $AP_{50}$ of RPDet is relatively high, which is believed as a better metric in many applications~\cite{YOLOv3}.

\vspace{-.5em}
\section{Conclusion}
In this paper, we propose \textit{RepPoints}, a representation for object detection that models fine-grained localization information and identifies local areas significant for object classification. Based on \textit{RepPoints}, we develop an object detector called RPDet that achieves competitive object detection performance without the need of anchors. Learning richer and more natural object representations like \textit{RepPoints} is a direction that holds much promise for object detection.

{\small
\bibliographystyle{ieee_fullname}
\bibliography{reppoints}
}
\clearpage
\section*{Appendix}
\appendix
\renewcommand{\thesection}{A\arabic{section}}  

\section{Relationship between RepPoints and Deformable RoI pooling}
In this section, we explain the differences between our method and deformable RoI pooling [4] in greater detail. We first describe the translation sensitivity of the regression step in the object detection pipeline. Then, we discuss how deformable RoI pooling \cite{DCN} works and why it does not provide a geometric representation of objects, unlike the proposed RepPoints representation.

\paragraph{Translation Sensitivity}
We explain the translation sensitivity of the regression step in the context of bounding boxes. Denote a rectangular bounding box proposal before regression as $\mathcal{B}_P$ and the ground-truth bounding box as $\mathcal{B}_{GT}$. The target for bounding box regression can then be expressed as
\begin{equation}
    T_P=\mathcal{F}(\mathcal{B}_{P},\mathcal{B}_{GT}),
\end{equation}
where $\mathcal{F}$ is a function for transforming $\mathcal{B}_P$ to $\mathcal{B}_{GT}$. This transformation is conventionally learned as a regression function $\mathcal{R}_B$:
\begin{equation}
    \mathcal{R}_B(\mathcal{P}_B(I, \mathcal{B}_P))=T_P=\mathcal{F}(\mathcal{B}_{P},\mathcal{B}_{GT}),
\end{equation}
where $I$ is the input image and $\mathcal{P}_B$ is a pooling function defined over the rectangular proposal, e.g., direct cropping of the image \cite{girshick2014rich}, RoIPooling \cite{ren2015faster}, or RoIAlign \cite{Mask-rcnn}. This formulation aims to predict the relative displacement to the ground truth box based on features within the area of $\mathcal{B}_P$. Shifts in $\mathcal{B}_P$ should change the target accordingly:
\begin{equation}
    \mathcal{R}_B(\mathcal{P}_B(I, \mathcal{B}_P+\Delta \mathcal{B}))=\mathcal{F}(\mathcal{B}_{P}+\Delta \mathcal{B},\mathcal{B}_{GT}).
\end{equation}

Thus, the pooled feature $\mathcal{P}_B(I, B_P)$ should be sensitive to the box proposal $B_P$. Specifically, for any pair of proposals $\mathcal{B}_1 \ne \mathcal{B}_2$, we should have $\mathcal{P}_B(I, B_1) \ne \mathcal{P}_B(I, B_2)$. Most existing feature extractors $\mathcal{P}_B$ satisfy this property. Note that the improvement of RoIAlign \cite{Mask-rcnn} over RoIPooling \cite{ren2015faster} is partly due to this guaranteed translation sensitivity.

\begin{figure}
    \centering
    \includegraphics[width=\linewidth]{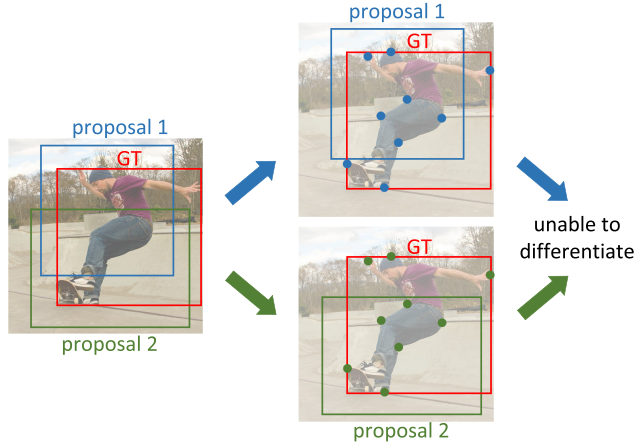}
\caption{Illustration that deformable RoI pooling \cite{DCN} is unable to serve as a geometric object representation, as discussed in Section 4 in the main paper. We consider two bounding box regressions based on different proposals. Assume that deformable RoI pooling \cite{DCN} can learn a similar geometric object representation where the two sets of sample points lie at similar locations over the object of interest. For that to happen, the sampled features would need to be similar, such that the two proposals cannot be differentiated. However, deformable RoI pooling
\cite{DCN} can indeed differentiate nearby object proposals, leading to a contradiction. Thus, it is concluded that deformable RoI pooling \cite{DCN} cannot learn the geometric representation of objects. }
    \label{fig::dpool_bug}
\end{figure}

\begin{figure}
    \centering
    \includegraphics[width=\linewidth]{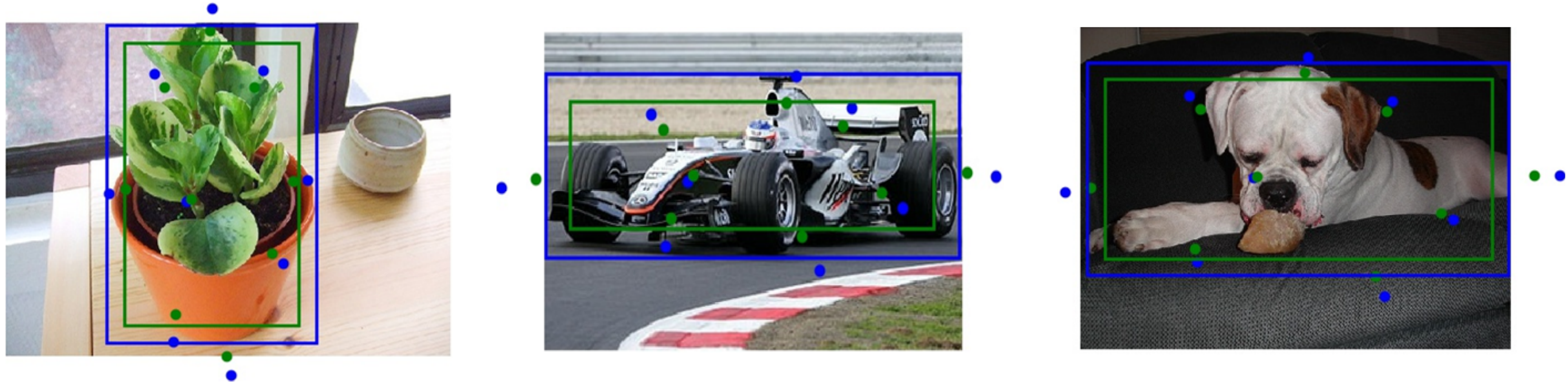}
    \caption{Visualization of the learned sample points of 3$\times$3 deformable RoI pooling \cite{DCN}. It is shown that the scale of sample points changes as the scale of the proposal changes, indicating that the sample points do not adapt to form a geometric \emph{object} representation.}
    \label{fig::dpool_demo}
\end{figure}

\paragraph{Analysis of Deformable RoI Pooling.}
For deformable RoI pooling \cite{DCN}, the system generates a pointwise deformation of samples on a regular grid \cite{Mask-rcnn} to produce a set of sample points $S_P$ for each proposal. This can be formulated as
\begin{equation}
    S_P=\mathcal{D}(I, \mathcal{B}_P),
\end{equation}
where $\mathcal{D}$ is the function for generating the sample points. Then, bounding box regression aims to learn a regression function $\mathcal{R}_S$ which utilizes the sampled features via $S_P$ to predict the target $T_P$ as follows:
\begin{equation}
    \mathcal{R}_S(\mathcal{P_S}(I,S_P))=T_P=\mathcal{F}(\mathcal{B}_{P},\mathcal{B}_{GT})
    \label{eq::temp5}
\end{equation}
where $\mathcal P_S$ is the pooling function with respect to the sample points $S_P$. 

From the translation sensitivity property, we have $\mathcal{P}_S(I, \mathcal{D}(I, \mathcal{B}_1)) \ne \mathcal{P}_S(I, \mathcal{D}(I, \mathcal{B}_2)), \forall \mathcal{B}_1 \ne \mathcal{B}_2$. Because the pooled feature $\mathcal{P}_S(I, \mathcal{D}(I, \mathcal{B}))$ is determined by the locations of sample points $\mathcal{D}(I, \mathcal{B})$, we have $\mathcal{D}(I, \mathcal{B}_1) \ne \mathcal{D}(I, \mathcal{B}_2), \forall \mathcal{B}_1 \ne \mathcal{B}_2$. This means that for two different proposals $\mathcal{B}_1$ and $\mathcal{B}_2$ of the same object, the sample points of these two proposals by deformable RoI pooling should be different. Hence, the sample points of different proposals cannot correspond to the geometry of the same object. They represent a property of the proposals rather than the geometry of the object.

Figure \ref{fig::dpool_bug} illustrates the contradiction that arises if deformable RoI pooling were a representation of object geometry. Moreover, Figure \ref{fig::dpool_demo} illustrates that, for the learned sample points of two proposals for the same object by deformable RoI pooling, the sample points represent a property of the proposals instead of the geometry of the object. 

\paragraph{RepPoints}
In contrast to deformable RoI pooling where the pooled features represent the original bounding box proposals, the features extracted from RepPoints localize the object. As it is not restricted by translation sensitivity requirements, RepPoints can learn a geometric representation of \emph{objects} when localization supervision on the corresponding pseudo box is provided (see Figure 4 in the main paper). While object localization supervision is not applied on the sample points of deformable RoI pooling, we show in Table 2 in the main paper that such supervision is crucial for RepPoints.

It is worth noting that deformable RoI pooling \cite{DCN} is shown to be complementary to the RepPoints representation (see Table 6 in the main paper), further indicating their different functionality.

\section{More Benchmark Results for RPDet}
\begin{table}[tb]
	\begin{center}
	\begin{tabular}{c|c|c|c|c}
	\hline
	method & backbone & ms train & ms test & AP \\
	\hline
	\hline
	RPDet & R-50 & & & 38.6 \\
	\hline
	& R-50 & \checkmark & & 40.8 \\
	\hline
	& R-50 & \checkmark & \checkmark & 42.2 \\
	\hline
	\hline
	& R-101 & & & 40.3 \\
	\hline
	& R-101 & \checkmark & & 42.3 \\
	\hline
	& R-101 & \checkmark & \checkmark & 44.1 \\
	\hline
	\hline
	& R-101-DCN & & & 43.0 \\
	\hline
	& R-101-DCN & \checkmark & & 44.8 \\
	\hline
	& R-101-DCN & \checkmark & \checkmark & 46.4 \\
	\hline
	\hline
	& X-101-DCN & & & 44.5 \\
	\hline
	& X-101-DCN & \checkmark & & 45.6 \\
	\hline
	& X-101-DCN & \checkmark & \checkmark & 46.8 \\
	\hline
	\end{tabular}
	\end{center}
\caption{Benchmark results of RPDet on MS-COCO \cite{MSCOCO} validation set (\texttt{minival}). All the models here are trained with FPN \cite{FPN} under the `2x' setting \cite{Detectron2018}. For the backbone notation, `R-50' and `R-101' denotes ResNet-50 and ResNet-101 \cite{ResNet} respectively. `R-101-DCN' denotes ResNet-101 with all convolution layers substituted with deformable convolution layers \cite{DCN}. `X' denotes the ResNeXt-101 \cite{ResNeXt} backbone. ``ms'' indicates multi-scale. }
\label{tab::benchmark}
\end{table}

We present more benchmark results of our proposed detector RPDet in Table \ref{tab::benchmark}. Our PyTorch implementation is available at \href{https://github.com/microsoft/RepPoints}{https://github.com/microsoft/RepPoints}. All models were tested on MS-COCO \cite{MSCOCO} validation set (\texttt{minival}). 

\noindent \emph{Multi-scale training and test settings.} In multi-scale training, for each mini-batch, the shorter side is randomly selected from a range of $[480, 960]$. In multi-scale testing, we first resize each image to a shorter side of $\{400, 600, 800, 1000, 1200, 1400\}$. Then the detection results (before NMS) from all scales are merged, followed by a NMS step to produce the final detection results.

\end{document}